\pdfoutput=1
\documentclass[10pt,twocolumn,letterpaper]{article}

\usepackage{cvpr}
\usepackage{times}
\usepackage{epsfig}
\usepackage{graphicx}
\usepackage{amsmath}
\usepackage{amssymb}
\usepackage{adjustbox}
\usepackage{commath}
\usepackage{comment}
\usepackage[tracking=true]{microtype}

\newcommand{\powerset}{\raisebox{.15\baselineskip}{\Large\ensuremath{\wp}}}

\usepackage[breaklinks=true,bookmarks=false]{hyperref}

\cvprfinalcopy % *** Uncomment this line for the final submission

\def\cvprPaperID{****} % *** Enter the CVPR Paper ID here

\begin{document}

%%%%%%%%% TITLE
\title{Partial Domain Adaptation Using Selective Representation Learning For Class-Weight Computation}

\author{Sandipan Choudhuri*, Riti Paul*, Arunabha Sen, Baoxin Li, Hemanth Venkateswara\\
CIDSE\\
Arizona State University\\
{\tt\small \{s.choudhuri, rpaul12, asen, Baoxin.Li, hkdv1\}@asu.edu}
}

\maketitle
%\thispagestyle{empty}

%%%%%%%%% ABSTRACT
\begin{abstract}

The generalization power of deep-learning models is dependent on rich-labelled data. This supervision using large-scaled annotated information is restrictive in most real-world scenarios where data collection and their annotation involve huge cost. Various domain adaptation techniques exist in literature that bridge this distribution discrepancy. However, a majority of these models require the label sets of both the domains to be identical. To tackle a more practical and challenging scenario, we formulate the problem statement from a partial domain adaptation perspective, where the source label set is a super set of the target label set. Driven by the motivation that image styles are private to each domain, in this work, we develop a method that identifies outlier classes exclusively from image content information and train a label classifier exclusively on class-content from source images. Additionally, elimination of negative transfer of samples from classes private to the source domain is achieved by transforming the soft class-level weights into two clusters, 0 (outlier source classes) and 1 (shared classes) by maximizing the between-cluster variance between them.

\begin{comment}
Standard domain adaptation techniques require label sets of both domains to be identical. To tackle a more practical scenario, we formulate the problem from a partial domain adaptation perspective; identifying outlier source classes exclusively from image content information. Driven by the fact that image styles are private to each domain, a label classifier is trained exclusively on class-content information from source domain using adversarial learning. Additionally, elimination of negative transfer of samples from classes private to source domain is achieved by partitioning soft class-level weights into outlier and shared classes through maximizing between-cluster variance.
\end{comment}

\end{abstract}

%%%%%%%%% BODY TEXT
\section{Introduction}

%\narrowstyle

Deep neural networks have remarkably leveraged the performance of a varied spectrum of models aimed at catering different machine learning problems. The generalization power of such models is, however, contingent on the availability of large-scale annotated data. This supervision using rich-labelled data 
is restrictive in some real-world applications where data collection and its annotation incur huge expenses. To circumvent this rich labelling procedure, techniques that utilize label information and knowledge from a related domain can be employed. However, the distribution shift between the datasets representing different domains poses a major bottleneck when designing networks for adapting to new tasks on unlabelled data. A considerable proportion of domain adaptation techniques exist in literature that bridges this distribution discrepancy by learning domain invariant representations of the data from two different domains. The models, thus constructed, can be directly deployed to unlabelled domain data. Although, these models form ideal candidates for solving the task at hand, a majority of them deals with a naive scenario where the label sets of the two domains (labeled source and unlabeled target domains) are equal. The setup turns out to be more realistic if the restriction on the label set overlap is relaxed. To illustrate further, in the era of big data, it is not difficult to envision transferring of knowledge from a large-scale source dataset, containing information from a wide variety of classes, to a smaller unlabelled target dataset, where it is safe to assume that the class label set of the smaller dataset is already contained within that of the larger dataset. 

Prior works on partial transfer learning \cite{cao2018partial,zhang2018importance,cao2018partial2,cao2019learning} have attempted to find representations that are shared between the source and target domains that circumvent negative transfer by penalizing the inclusion of outlier source classes. However, the feature extractors present in such techniques jointly process both the styles and content components of the domain representations, which are further processed for outlier class computation. Such style components act as noise and contaminates the initial phases of training, thereby leading to erroneous computation of outlier classes. In this work, we propose to eliminate that problem by imposing domain discrimination exclusively on the class-content representation. Current methods, focus on eliminating outlier source classes by introducing soft class weights, which means there lies possibilities of some degrees of transfer from samples in the outlier classes. Our technique circumvents this issue by introducing a thresholding mechanism by maximizing the between-cluster variances that binarizes the quantification of transferability of samples from the source domain. 

To summarize, we propose a methodology that allows to produce split representations of class-content and image style. Since image styles are private to each domain, the classifier is trained exclusively on class-content information common between both the domains. Furthermore, to eliminate the negative transfer of samples from classes private to source domain, we introduce hard class weights by transforming the soft-class weights into two categories (0 and 1) by maximizing between-cluster variance between them.

\section{Problem Formulation}

Equivalent to a standard transfer learning scenario, in this setup we are furnished with information from two different domains, source $S$ and target $T$ respectively. $D_s=\{(x^i_s,y^i_s)\}_{i=1}^{n_s}$ is a dataset of $n_s$ samples, representing domain $S$, where every data point $x^i_s \in \mathbb{R}^{d}$ is drawn in an i.i.d. fashion from a distribution $p_s$ and is associated with a label $y^i_s \in C_s$. An unlabelled dataset $D_t=\{x^i_t\}_{i=1}^{n_t}$, representing domain $T$, consists of $n_t$ data points where each $x^i_t \in \mathbb{R}^{d}$ is sampled in an i.i.d. manner from a different distribution $p_t$ ($p_t \neq p_s$). The goal of this paper is to design a classifier hypothesis $f_c: f_c(x_t) \rightarrow y_t$ for data $(x_t,y_t)$ in $T$, $y_t \in C_t$, that minimizes the target classification risk $r_t = P_{x_t \sim p_t} [f_c(x_t) \neq y_t]$. However, since the class label information $y_t$ is unavailable during hypothesis learning, the task is performed by leveraging Source domain supervision.

%This work deals with a scenario more general and challenging than the standard Closed-Set Domain Adaptation problem, namely Partial Domain Adaptation (PDA). It relaxes the fully overlapping label space assumption between $S$ and $T$ by allowing the label set $C_t$ of the Target dataset $D_t$ to be a subset of the source data label space $C_s$, i.e. $C_t \subseteq C_s$. 

%In standard domain adaptation techniques, one of the main difficulties is that the target domain has no labeled data and thus the classifier trained on
%source domain $D_{s}$ cannot be directly applied to target domain $D_{t}$, due to the distribution discrepancy of $p_t \neq p_s$. 

In partial domain adaptation, in addition to the existing distribution discrepancy, the model faces the challenge of transferring relevant data from source to target as the knowledge of shared labels between source and target label space is unknown ($C_{t}$ is unavailable during training). Therefore, as a crucial step to obtain accurate classifications, it is essential to prevent learning from data samples associated with labels private to $S$ i.e., label set represented by $C_s$\textbackslash$C_t$ ($C_s$\textbackslash$C_t = C_s-C_t$). Since information on $C_t$ is concealed during training, identification of the shared label set $C_t$ among a pool of $2^{|C_s|}$ possible label sets in the power set $\powerset(C_s)$ that minimizes $r_t$ is a non-trivial task.

\begin{figure*}[h]
    \centering
    \includegraphics[width = .85\linewidth, height = 11.5 cm]{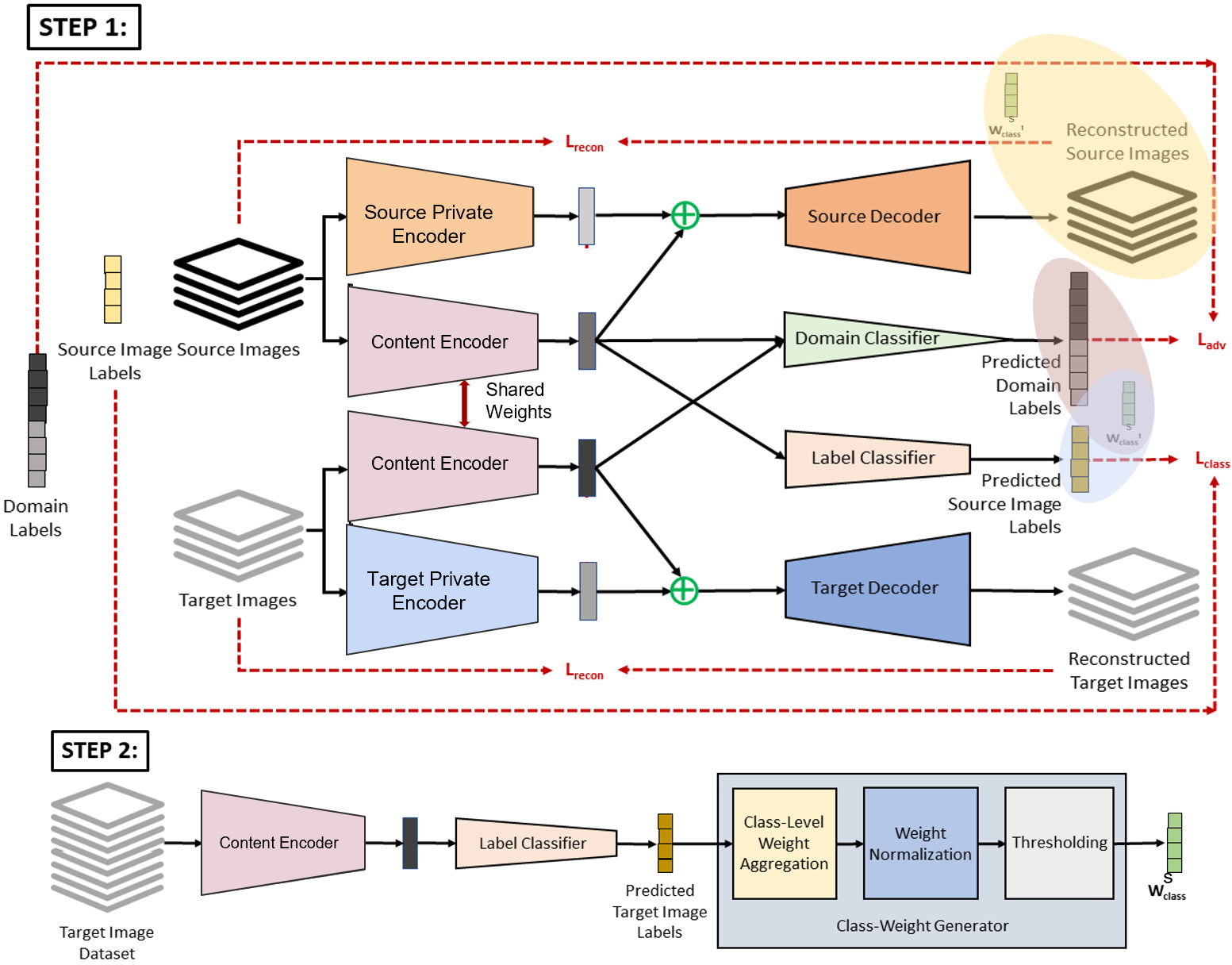}
    \caption{ A schematic diagram of the proposed network. In STEP 1, The source and target content encoder (both the encoders share weights) learns to generate representations that represent content relevant for classification. The common content representation learning is achieved with the aid of the domain classifier using adversarial learning and a label classifier. The source and target style encoders learn features private to images in their respective domains. The source and target decoders accept representations containing common content and style information private to each domain and tries to reconstruct the input images. The loss $L_{diff}$ ensures that representations learnt by the content and style encoders are different. STEP 2 covers computation of class weights using output scores of the label classifier on the target images. In this module, the threshold technique described in eq. \ref{threshold}, is utilized to binarize the soft labels into hard labels, thereby completely restricting negative transfer from outlier class samples in the source domain. }    
    \label{label2}
\end{figure*}

\section{Proposed Method}
%Given a labeled set of source samples and unlabeled set of target sample, our primary goal is to train a classifier on data from the source domain that generalizes very well on the target domain. An additional objective of this method is to identify the outlier classes from the source domain and downplay their role played in the negative transfer during hypothesis learning. 

The proposed network, inspired by \cite{bousmalis2016domain}, is designed to model the private and shared representations of the different domains explicitly. The private representations are specific to each domain and the shared representations are common between domains. To model this property, we use three separate set of encoders. Two private encoders are trained to capture the domain specific features. The shared encoder is trained to capture features that are common across domains and is trained on both the labeled source and unlabeled target samples. A variety of loss functions are utilized in the model to capture different features relevant to the task at hand. Furthermore, to ensure that the content of the private representations are still useful and to generalize even better, we apply image reconstructions over the shared and private representations using source and target decoders. A classifier is trained on the shared representations to improve the generalization across domains and avoid being influenced by factors specific to each domain. The loss functions are defined as follows:
\begin{multline}
    L_{recon} = \\\sum\limits_{x_i^s \in D_s}W_{class_{y^s_i}}^s L_{sim}(DE_s(E_{sh}(x_i^s) \oplus E_{s}(x_i^s)), \hat{x_i^s})\\ + \sum\limits_{x_i^t \in D_t}L_{sim}(DE_t(E_{sh}(x_i^t) \oplus E_{t}(x_i^t)), \hat{x_i^t}) \tag{1}
    \label{equation1}
\end{multline}
where $L_{sim}$ is defined as:
\begin{equation*}
    L_{sim}(x, \hat{x}) = \frac{1}{k}\norm{x - \hat{x}}^2_2 + \frac{1}{k^2}(\abs{x - \hat{x}}.1_k)^2     \tag{2}
    \label{equation2}
\end{equation*}
$L_{recon}$ is the scale-invariant version of mean squared error term which penalizes differences between pairs of pixels. This allows the model to learn to reproduce the overall shape of the objects being modeled, without expending modeling power on the absolute color or intensity of the inputs. 

\begin{comment}
The difference loss is used to train the private and shared encoders to extract different aspects of the input. We use orthogonality constraint between shared and private representations of each domain to enable the encoders to learn distinct features of the input. Let $H^s_c$ and $H^t_c$ be matrices whose columns are the hidden shared representations $h^s_c = E_{sh}(x^s)$ and $h^t_c = E_{sh}(x^t)$ from samples of source and target data respectively. Similarly, let $H^s_p$ and $H^t_p$ be matrices whose columns are the private representation $h^s_p = E^s_p(x^s)$ and $h^t_p = E^t_p(x^t)$ from samples of source and target data respectively. The difference loss, which encourages orthogonality between the shared and the private representations of each domain, is represented as follows:
\begin{equation*}
    L_{diff} = \norm{H_c^s^TH^s_p}^2_F + \norm{H_c^t^TH^t_p}^2_F    \tag{3}
\end{equation*}
where $\norm{.}^2_F$ is the squared Frobenius form. 
\end{comment}

In the proposed model, we have a label classifier $G_y$ and a domain classifier network $G_d$. Our framework aims to reduce the shift of shared classes between source and target domains. The loss of the label classifier is as follows:
\begin{equation*}
    L_{class} = \frac{1}{n_s}  \sum\limits_{x_i^s \in D_s} W_{class_{y^s_i}}^s L(G_y(E_{sh}(x_i^s)), y_i)  \tag{4}
\end{equation*}
where $E_{sh}$ provides the shared the representation for each sample, with $G_y$ trained using the cross-entropy loss $L$. $G_d$, on the other hand, is learned by minimizing the following:
\begin{multline}
    L_{adv} = -\frac{1}{n_s}\sum\limits_{x_i^s \in D_s} W_{class_{y^s_i}}^s[ logG_d(E_{sh}(x_i^s))]\\ -\frac{1}{n_t}\sum\limits_{x_i^t \in D_t}[1 - logG_d(E_{sh}(x_i^t))]      \tag{5}
    \label{L_adv}
\end{multline}

Entropy minimization regularization is generally used to reduce the adverse-effects caused by classifier uncertainty, due to large domain shift and difficulty in transferring samples. In the proposed model, we make use of the entropy minimization  principle, which is defined as follows:  

\begin{equation*}
    L_{ent} = - \frac{1}{n_t} \sum\limits_{i=1}^{n_t} \sum\limits_{c=1}^{|C_s|} \hat{y}^t_{i_c} y^t_{i_c}
\end{equation*}

It is important to address the transferability of shared source classes. To find out which class in source domain belongs to the shared classes, we make use of shared class weights $W_{class}^s$, which is a $|C_s|$ dimensional vector; each element $w_j^s$ represents the probability of the $j^{th}$ class of the source domain belonging to the shared classes. $W^s_{class}$ is defined as following:
\begin{equation*}
    W_{class}^s = \frac{1}{n_t}\Sigma_{i=1}^{n_t}G_y(E_{sh}(x_i^t))    \tag{6}
\end{equation*}

A recent work \cite{hu2019multi} has utilized the concept of class weights. However, they are often computed upon feature descriptors that may contain both content and style of domain images. In our proposed model, we utilize domain separation networks to capture different representation components of any given input sample. The common and the domain specific features are separated using the different sets of encoders. The shared feature representations, thus obtained, are used to train the domain classifier, whose loss ($L_{adv}$) when back-propagated aids in generating more robust domain-invariant features. Instead of using features that contain both style and content of the respective domain, we are proposing the use of features shared across the domains in computing the shared class weights. Our conjecture is that this will improve the generalization across domains and avoid being influenced by domain specific factors. 
%The work in \cite{} shows that using hard labels is beneficial for selecting shared source samples. To be added - hard label algorithm and shared sample classifier. 

\begin{table*}[htbp!]
\begin{adjustbox}{width=1\textwidth}

 \begin{tabular}{c | c c c c c c c c c c c c | c} 
 \hline
 Method & Ar $\rightarrow$ Cl & Ar $\rightarrow$ Pr & Ar $\rightarrow$ Rw & Cl $\rightarrow$ Ar & Cl $\rightarrow$ Pr & Cl $\rightarrow$ Rw & Pr $\rightarrow$ Ar & Pr $\rightarrow$ Cl & Pr $\rightarrow$ Rw & Rw $\rightarrow$ Ar & Rw $\rightarrow$ Cl & Rw $\rightarrow$ Pr & Avg. \\ 
 \hline
Resnet-50\cite{he2016deep} & 46.33 & 67.51 & 75.87 & 59.14 & 59.94 & 62.73 & 58.22 & 41.79 & 74.88 & 67.40 & 48.18 & 74.17 & 61.35\\
DAN\cite{long2015learning} & 43.76 & 67.90 & 77.47 & 63.73 & 58.99 & 67.59 & 56.84 & 37.07 & 76.37 & 69.15 & 44.30 & 77.48 & 61.72\\
DANN\cite{ganin2016domain} & 45.23 & 68.79 & 79.21 & 64.56 & 60.01 & 68.29 & 57.56 & 38.89 & 77.45 & 70.28 & 45.23 & 78.32 & 62.82\\
ADDA\cite{tzeng2017adversarial} & 45.23 & 68.79 & 79.21 & 64.56 & 60.01 & 68.29 & 57.56 & 38.89 & 77.45 & 70.28 & 45.23 & 78.32 & 62.82\\
RTN\cite{long2016unsupervised} & 49.31 & 57.70 & 80.07 & 63.54 & 63.47 & 73.38 & 65.11 & 41.73 & 75.32 & 63.18 & 43.57 & 80.50 & 63.07\\
IWAN\cite{zhang2018importance} & 53.94 & 54.45 & 78.12 & 61.31 & 47.95 & 63.32 & 54.17 & 52.02 & 81.28 & 76.46 & 56.75 & 82.90 & 63.56\\
SAN\cite{cao2018partial} & 44.42 & 68.68 & 74.60 & \textbf{67.49} & \textbf{64.99} & \textbf{77.80} & 59.78 & 44.72 & 80.07 & 72.18 & 50.21 & 78.66 & 65.30\\
PADA\cite{cao2018partial2} & 51.95 & 67.00 & 78.74 & 52.16 & 53.78 & 59.03 & 52.61 & 43.22 & 78.79 & 73.73 & 56.60 & 77.09 & 62.06\\
SSPDA\cite{cao2019learning} & 52.02 & 63.64 & 77.95 & 65.66 & 59.31 & 73.48 & 70.49 & 51.54 & \textbf{84.89} & 76.25 & \textbf{60.74} & 80.86 & 68.07\\

 \hline
 
Our approach & \textbf{56.21} & \textbf{73.34} & \textbf{80.63} & 64.08 & 61.72 & 66.41 & \textbf{70.83} & \textbf{53.13} & 83.57 & \textbf{77.01} & 58.31 & \textbf{81.24} & \textbf{68.87} \\ [1ex] 
 \hline
\end{tabular}
\end{adjustbox}
\caption{Classification accuracy (\%) of Partial Domain Adaptation on Office-Home dataset with Resnet-50 as backbone.}
\end{table*}

\begin{table*}[htbp!]
\centering
\begin{adjustbox}{width=0.55\textwidth}
 \begin{tabular}{c | c c c c c c | c} 
 \hline
 
Method & A $\rightarrow$ W & A $\rightarrow$ D &  W $\rightarrow$ A & W $\rightarrow$ D & D $\rightarrow$ A & D $\rightarrow$ W & Avg.\\
\hline

Resnet-50\cite{he2016deep} & 75.59 & 83.44 & 84.97 & 98.09 & 83.92 & 96.27 & 87.05\\
DAN\cite{long2015learning} & 59.32 & 61.78 & 67.64 & 90.45 & 74.95 & 73.90 & 71.34\\
DANN\cite{ganin2016domain} & 73.56 & 81.53 & 86.12 & 98.73 & 82.78 & 96.27 & 86.50\\
ADDA\cite{tzeng2017adversarial} & 75.67 & 83.41 & 84.25 & 99.85 & 83.62 & 95.38 & 87.03\\
RTN\cite{long2016unsupervised} & 78.98 & 77.07 & 89.46 & 85.35 & 89.25 & 93.22 & 85.56\\
IWAN\cite{zhang2018importance} & 89.15 & 90.45 & 94.26 & 99.36 & \textbf{95.62} & \textbf{99.32} & 94.69\\
SAN\cite{cao2018partial} & 90.90 & 94.27 & 88.73 & 99.36 & 94.15 & \textbf{99.32} & \textbf{94.96}\\
PADA\cite{cao2018partial2} & 86.54 & 82.17 & \textbf{95.41} & \textbf{100.00} & 92.69 & \textbf{99.32} & 92.69\\
SSPDA\cite{cao2019learning} & 91.52 & 90.87 & 94.36 & 98.94 & 90.61 & 92.88 & 93.20\\

 \hline
Our approach & \textbf{92.07} & \textbf{94.46} & 93.72 & 99.24 & 93.68 & 95.84 & 94.84\\
\hline
\end{tabular}

\end{adjustbox}
\caption{Classification accuracy (\%) of Partial Domain Adaptation on Office-31 dataset with Resnet-50 as backbone.}
\end{table*}

Current methods \cite{cao2018partial,zhang2018importance,cao2018partial2}, identify the outlier source classes by introducing soft class weights. This signifies that there exists possibilities of some degrees of transfer from samples in the outlier classes. Our technique eliminates this issue by introducing a thresholding mechanism by maximizing the between-cluster variances that binarizes the quantification of transferability of samples from the source domain. We divide these source classes into two groups : $C_{out}$ and $C_{share}$. $C_{out}$ denotes classes with weights $\in [0,t)$ and $C_{share}$ denotes classes with weights $[t,1]$. The probability of class occurrences $w_{out}$ and $w_{share}$ and the class mean weights $\mu_{out}$ and $\mu_{share}$ are defined as follows:

\begin{equation*}
    w_{out} = \frac{|c_{out}|}{|C_s|}   , \mu_{out} = \sum_{w^s_{{class}_j}=0}^t \frac{w^s_{{class}_j}}{|c_{out}|}   \tag{7}
\end{equation*}

\vspace{-7mm}

\begin{equation*}
    w_{share} = \frac{|c_{share}|}{|C_s|} , \mu_{share} = \sum_{w^s_{{class}_j}=t}^1 \frac{w^s_{{class}_j}}{|c_{share}|}  \tag{8}
\end{equation*}

\vspace{-5mm}

\begin{equation*}
\mu_{total} = \mu_{out} * w_{out} + \mu_{share} * w_{share}     \tag{9}
\end{equation*}

This is followed by measuring the between-cluster variance 

\begin{equation*}
 \delta^2 = (\mu_{out}-\mu_{total})^2 * w_{out} + (\mu_{share}-\mu_{total})^2 * w_{share}      \tag{10}
\end{equation*}

A larger value of $\delta^2$ signifies greater differences between the two clusters. Consequently, the value of $t$ producing the largest variance value $\delta^2$ is the ideal candidate, i.e., 

\begin{equation*}
    t = argmax(\delta^2)    \tag{11}
    \label{threshold}
\end{equation*}

When the value of threshold $t$ is maximized, it implies that the probability of misclassification of $C_{out}$ and $C_{share}$ is minimized. By utilizing this threshold value, we binarize the soft class weight vector $W_{class}^s$, where values in the range $[0,t) = 0$ and that within $[t,1] = 1$. 

To sum up, the overall loss function is as follows ($\lambda$: regularization parameter for image reconstruction loss):

\begin{equation}
L = \lambda L_{recon}+L_{class}+L_{adv}+L_{ent}
\end{equation}

\section{Experiments}

To evaluate the efficacy of the proposed approach, we perform experiments on two benchmark datasets (Office-Home\cite{venkateswara2017deep} and Office-31\cite{saenko2010adapting}) across multiple tasks. The following sections highlight the datasets, tasks for experimentation and the network hyper-parameters.

\subsection{Datasets} For performance evaluation of the proposed method, we utilize two commonly used datasets for domain adaptation, namely Office-Home and Office-31.
The relatively small Office-31\cite{saenko2010adapting} dataset contains 4652 images from 31 different classes with images from three domains: Webcam (W), Amazon (A) and DSLR (D). Following the method in \cite{cao2018partial2}, we build the target dataset with images from 10 different categories. The evaluation is carried out on 6 different tasks, namely A$\rightarrow$W, A$\rightarrow$D, W$\rightarrow$A, W$\rightarrow$D, D$\rightarrow$A and D$\rightarrow$W. 

To further establish the efficacy of our model, we tested it on the larger Office-Home \cite{venkateswara2017deep} dataset (a collection of around 15,500 images), with images collected from four different domains: Real-world (Rw), Artistic (Ar), Product (Pr) and Clip Art (Cl). Similar to the procedure utilized in \cite{cao2018partial2}, we build the target dataset from 25 classes and the source domain with images from 65 classes. 12 different domain adaptation tasks were arranged for evaluation, namely: Ar$\rightarrow$Cl, Ar$\rightarrow$Pr, Ar$\rightarrow$Rw, Cl$\rightarrow$Ar, Cl$\rightarrow$Pr, Cl$\rightarrow$Rw, Pr$\rightarrow$Ar, Pr$\rightarrow$Cl, Pr$\rightarrow$Rw, Rw$\rightarrow$Ar, Rw$\rightarrow$Cl and Rw$\rightarrow$Pr.

\subsection{Comparison Models}
The performance of the proposed model is compared with the state-of-the-art deep learning models addressing partial domain adaptation: Resnet-50 \cite{he2016deep}, Deep Adaptation Network (DAN) \cite{long2015learning}, Domain Adversarial Neural Network (DANN) \cite{ganin2016domain}, Adversarial Discriminative Domain Adaptation (ADDA) network \cite{tzeng2017adversarial}, Residual Transfer Networks (RTN) \cite{long2016unsupervised}, Importance Weighted Adversarial Nets (IWAN) \cite{zhang2018importance}, Selective Adversarial Network (SAN) \cite{cao2018partial}, Partial Adversarial Domain Adaptation(PADA) \cite{cao2018partial2} and class Subset Selection for Partial Domain Adaptation (SSPDA) \cite{cao2019learning}.

\subsection{Network Parameters}

We implement both the encoders using ResNet-50 \cite{he2016deep} and introduce a bottleneck layer of length 256 before the fully connected layers, as in DANN \cite{ganin2016domain}. The decoders are created using a series of three $3\times 3$ convolution (with \textit{relu} activation) and up-sampling layers, followed by a final convolutional layer. The new layers, introduced in the network, are trained from scratch with a learning rate 10 times faster than that of the fine-tuned layers. Mini-batch stochastic gradient descent(SGD) is utilized, with momentum set to 0.9 and learning rate strategy in DANN \cite{ganin2016domain}. The loss weight $lambda$ for image reconstruction is set to $10^{-4}$. The transferability of shared source classes is addressed by using binarized aggregated output scores of the label classifier on target images. 

\section{Results}

From Tables 1 and 2, it is observed that approaches specifically targeted towards mitigating distribution in a partial domain adaptation setup yield better accuracy than standard domain adaptation methods like DAN \cite{long2015learning}, DANN \cite{ganin2016domain}, ADDA \cite{tzeng2017adversarial} and RTN \cite{long2016unsupervised}. When tested on the Office-31 dataset, the proposed model achieves best performance in two out of six tasks. It produces the second-best average accuracy value (trailing by 0.12\%) when compared to other baselines. During evaluation on a much larger and complex dataset (Office-Home), it is observed that our model outperforms the rest in seven out of fourteen tasks, in addition to achieving the best average performance.

\section{Conclusion}
This paper presents a novel domain-invariant feature learning framework for partial domain adaptation. The proposed model learns domain-invariant features by eliminating that image style properties private to each domain, and utilizing the class discriminating properties existing between domains. Furthermore, the negative transferability of the private source classes is diminished by utilizing a weighted mechanism which operates by maximizing the between-cluster variance. From experiments conducted on two benchmark datasets, it is established that our approach successfully tackles the partial domain adaptation problem.

%{\small
\bibliographystyle{ieee}
%\bibliography{egbib}

%}

\end{document}